\documentclass{ifacconf}
\usepackage{graphicx}      
\usepackage{natbib}        
\usepackage{xcolor}
\usepackage{amsmath}
\usepackage{amssymb}  
\usepackage{url}                

\usepackage{subcaption}   
\usepackage{dblfloatfix}

\usepackage{listings}
\begin{document}
\begin{frontmatter}

\title{A Prototyping Framework for Distributed Control of Multi-Robot Systems} 

\author[First]{Junaid Ahmed Memon} 
\author[First]{Allan Andre Do Nascimento} 
\author[First]{Kostas Margellos} 
\author[First]{Antonis Papachristodoulou}

\address[First]{Department of Engineering Science, University of Oxford, UK (e-mail: \{junaid.memon, allan.adn, kostas.margellos, antonis\}@eng.ox.ac.uk)}

\thanks{JAM acknowledges support from the Oxford Pakistan Programme. AAdN, AP, and KM acknowledge support from MathWorks. AP was additionally supported by EPSRC grants EP/X017982/1, EP/Y014073/1, and UKRI2108. For Open Access, the authors have applied a CC BY licence to any Author Accepted Manuscript arising from this submission. }

\begin{abstract}
This paper presents a prototyping framework for distributed control of multi-robot systems, aimed at bridging theory and practical testing of distributed optimization algorithms. Using the Single Program, Multiple Data (SPMD) paradigm, the framework emulates distributed control on a single computer, with each core running the same algorithm using local states and neighbour-to-neighbour communication. We demonstrate the framework on a four-quadrotor position-swapping task using a non-cooperative game-theoretic distributed algorithm. Computational time and trajectory data are compared across the supported dynamics levels: a point-mass model, a high-fidelity quadrotor model, and an experimental hardware testbed using Crazyflie quadcopters. The results show that the framework provides a low-cost and accessible approach for validating distributed algorithms.
\end{abstract}


\end{frontmatter}

\section{Introduction}

Multi-robot systems are increasingly used in demanding tasks, such as environmental monitoring, inspection, search and rescue where scalability, performance, and autonomy are needed. Although distributed optimisation and coordination algorithms enable scalable decision making, their physical validation remains challenging. Bridging simulation and hardware gap therefore requires modular, affordable, and reproducible testbeds that capture realistic sensing, communication, and computation constraints \cite{mokhtarian2024survey}. 

Testing distributed algorithms on robot fleets faces two main challenges. First, although distributed architectures improve scalability, fault tolerance, and reliability over centralised ones \citep{jamshidpey2025centralization}, switching between architectures or adapting code across them is often difficult. Second, simple models, digital twins, and hardware controllers are typically developed separately, increasing effort and slowing prototyping. A unified framework that preserves the robot-fleet structure across all testing stages is therefore essential for reliable, safe, and stable deployment \citep{shakeri2019design}, while accelerating the design, build and test cycle.

\subsection{Related Work}
\vspace{-1mm}
Distributed optimisation algorithms are well studied for multi-agent systems \cite{notarstefano2019distributed,yang2019survey}. Despite advances ranging from ADMM-based formation control \citep{stomberg2023cooperative} to decentralised sequential quadratic programming for non-linear MPC \citep{stomberg2025decentralized}, real-time hardware implementation remains difficult due to iterative communication, timing constraints, and heterogeneous agent dynamics. Although simulation studies are abundant, only few support seamless hardware transition \cite{shorinwa2024distributed}.

For experimental validation, several multi-robot testbeds have been developed. Scaled transport platforms such as the University of Delaware scaled smart city \citep{stager_scaled_2018} demonstrate cooperative vehicle coordination in structured environments. Swarm platforms such as Kilobot \citep{rubenstein2014kilobot} and Robotarium \citep{pickem_robotarium_2017, wilson_robotarium_2020} provide low-cost distributed coordination environments, but are typically limited to a single class of robot dynamics, e.g. differential-drive ground robots, and rely on localization setups that are difficult or expensive to replicate. Duckietown \citep{paull2017duckietown} and the Cambridge RoboMaster testbeds \citep{blumenkamp_cambridge_2024} extend capability, but remain tied to specific hardware or sensing modalities, making adaptation to new robot classes or distributed optimisation routines non-trivial. For multi-UAV platforms, the systems in \cite{deng_indoor_2020} and the distributed hardware implementations in \cite{king_distributed_2004} are particularly notable.

The range of available platforms reflects growing interest in experimental validation of distributed systems. However, prior work often requires dedicated hardware clusters, complex networking, or high-cost setups, limiting accessibility for small-scale academic and teaching environments. Moreover, to the best of our knowledge, none provides integrated simulation and hardware testing within a single framework that eliminates multiple tools and reduces testing time for distributed multi-robot algorithms.

\subsection{Contributions}

To address these shortcomings, this work proposes a \emph{unified} parallel-computing framework for rapid prototyping of distributed control algorithms across three fidelity levels: simplified models, high-fidelity digital twins, and real hardware. The main contributions are:
\begin{enumerate}
    \item We provide a unified workflow for testing distributed algorithms on simple models, high-fidelity digital twins, and hardware from the same workstation, addressing multi-stage validation needs \citep{mokhtarian2024survey}.
    
    \item Unlike testbeds limited to a single robot morphology, our approach can be extended to support a \emph{variety} of dynamical model. This enables easy transitions between heterogeneous systems.
    
    \item The framework uses the Single Program, Multiple Data (SPMD) paradigm \citep{flynn_very_1966, foster_designing_nodate, gropp_using_2014} to emulate decentralised execution, with each worker acting as an independent agent using local states and neighbour communication. This enables scalable algorithm-in-the-loop testing on a single multi-core workstation, avoiding complex networking while reducing simulation time through parallel execution.
    
    \item We adapt and incorporate a Crazyflie model \citep{kimmodel,forster_cfmodel_2015} along with hardware APIs, enabling consistent testing with simple model, high-fidelity model, and physical UAVs. With this approach only the actuation block changes while the rest of the distributed controller remains identical, fulfilling modularity.
    
    
    \item We demonstrate the utility of the framework through a distributed position-swapping example and release the full code%
\footnote{\url{https://tinyurl.com/smart-city-dcf}}for reproducibility and further development.
    
\end{enumerate}

Overall, we provide an end-to-end prototyping pipeline that significantly reduces testing time, lowers hardware requirements, and improves reproducibility. 

\subsection{Article Organization}

The remainder of this paper is organized as follows. Section~\ref{sec:Prototyping Framework} presents the proposed framework, including its workflow and computational setup for multi-fidelity testing of distributed optimization algorithms. Section~\ref{sec:Method} describes the implementation, software and hardware setup, and the four-UAV position-swapping problem. Section~\ref{sec:Results} reports the results for the simple model, digital twin, and hardware platform. Finally, Section~\ref{sec:conclusion} concludes the paper and discusses future platform extensions.


\section{Prototyping Framework}
\label{sec:Prototyping Framework}

\begin{figure}[!ht]
\begin{center}
\includegraphics[width=8.4cm]{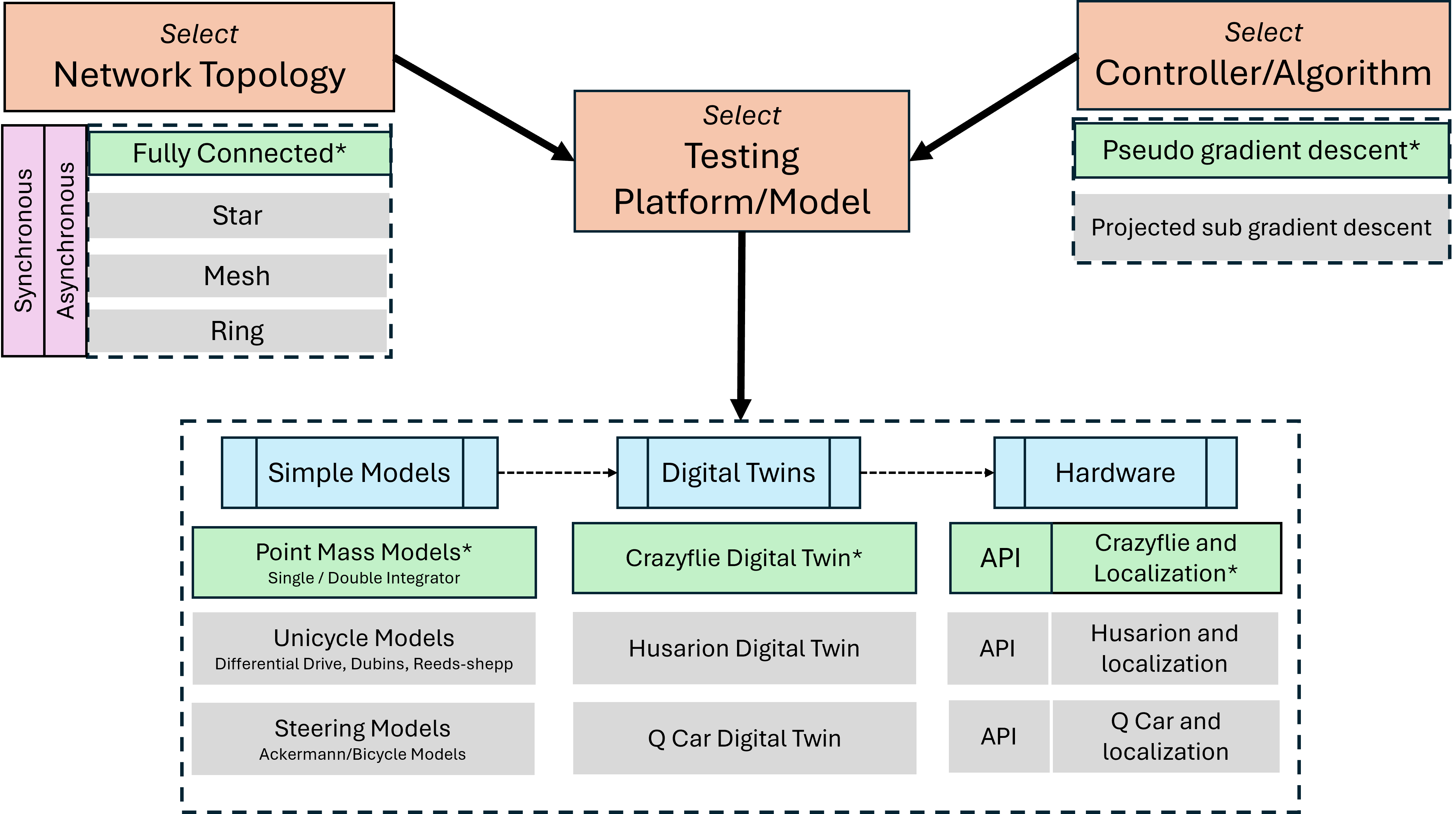}
\caption{Proposed workflow for testing distributed control on multi-robot systems.} 
\label{fig:workflow}
\end{center}
\end{figure}

We envision testing of distributed optimization algorithms with practical multi-robot deployments, by organizing the prototyping workflow into a structured sequence of decisions and execution stages, as illustrated in Figure~\ref{fig:workflow}.

The workflow involves selecting the network topology (e.g., fully connected, star, mesh, ring etc.), execution mode (synchronous or asynchronous), and a controller or algorithm such as distributed pseudo-gradient or projected subgradient descent. It further supports three testing levels: simple dynamical models for rapid prototyping, high-fidelity digital twins, and physical hardware with associated APIs and localization modules. The modular architecture enables deployment across different models and platforms through a common computational interface. In Fig.~\ref{fig:workflow}, currently supported features are shown in green (with *), while planned extensions are shown in grey.

\begin{figure}[!ht]
\begin{center}
\includegraphics[width=7.5cm]{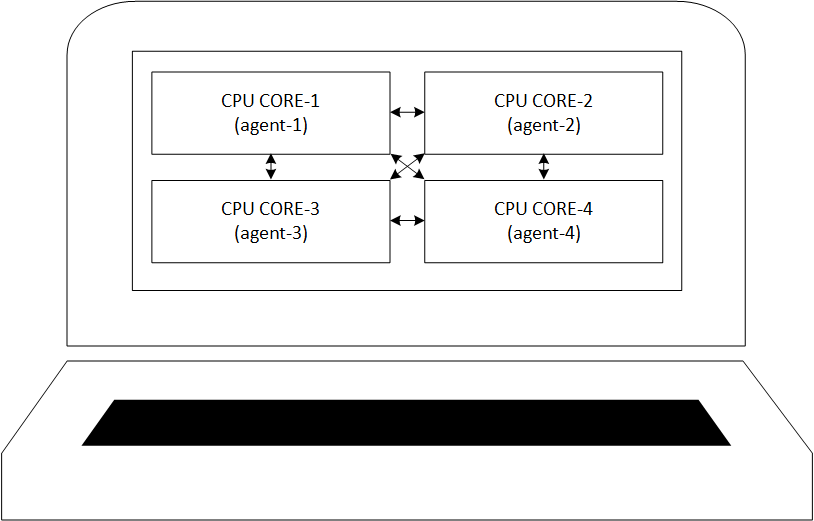}
\caption{Distributed control setup using a single computer.} 
\label{fig:distributed_setup}
\end{center}
\end{figure}

Figure~\ref{fig:workflow} also shows how these components integrate into a testing workflow at various levels of fidelity. First, any given algorithm should be tested on simple dynamical models (e.g., single or double integrators) to rapidly evaluate its performance and communication patterns. The second stage involves validating these algorithms with a digital twin (i.e. high-fidelity dynamic models) to capture underlying unmodeled dynamics (e.g., mechanics, aerodynamics), actuator limits, sensor noise, etc. Finally, the third stage transitions to real hardware, where the validated algorithms are deployed via the same algorithmic architecture, using an application layer to interface with multiple agents. All testing stages can be executed in a unified manner using a single multi-core workstation, with each core representing a logical agent in the multi-agent network, as shown in Figure~\ref{fig:distributed_setup}.


Framework employs the SPMD paradigm, where agents execute identical control programs using local states, decision variables, and neighbour information while exchanging only algorithm-required data. The approach reduces setup complexity and hardware cost while deploying parallel execution, logging, debugging, and integration with testing platforms. With this approach, we can emulate communication effects in code such as latency, packet loss, and bandwidth constraints for robustness evaluation.

\section{Methodology}
\label{sec:Method}

\subsection{Demonstration Problem}
\label{subsec:dem_example}
\vspace{-2mm}
For the framework demonstration, we consider a fleet of UAVs performing level flight while avoiding collisions and pursuing a shared objective. This yields an optimization problem coupled through the cost function (via the shared objective) and in the constraints (through collision avoidance), making it suitable for distributed computation.

\subsubsection{System model:}
\vspace{-2mm}
Consider a set $I = \{1,\ldots,N\}$, an indexed fleet of $N$ UAVs. For the \emph{control design} purposes, we assume that the UAVs share the same dynamics: $\ddot{p}^x_{i}(t) = a^x_{i}(t),\: \ddot{p}^y_{i}(t) = a^y_{i}(t)$, where $p^x_{i}(t), p^y_{i}(t)$ denotes the $i^{th}$-UAV's position in the $(x,y)$ plane respectively at time $t$. Similarly, we denote by $v^x_{i}(t), v^y_{i}(t), a^x_{i}(t), a^y_{i}(t)$ the $i^{th}$ UAV velocity and acceleration components in the $xy$-plane. Following a zero order hold discretization with sample time $t_s$, we obtain: 
\vspace{-2mm}

\begin{equation}
\label{cont_dyn}
z_i[k+1] =
\begin{bmatrix}
1 & 0 & t_s & 0\\
0 & 1 & 0 & t_s\\
0 & 0 & 1 & 0\\
0 & 0 & 0 & 1
\end{bmatrix}
z_i[k] +
\begin{bmatrix}
t_s^2/2 & 0\\
0 & t_s^2/2\\
t_s & 0\\
0 & t_s
\end{bmatrix}
a_i[k]
\end{equation}

 where $z_{i}[k]=[\: p^x_{i}[k] \: p^y_{i}[k] \: v^x_{i}[k] \: v^y_{i}[k]\:]^{\top} \in \mathbb{R}^4$ denotes the state, $a_{i}[k] = [ \:a^x_{i}[k] \: a^y_{i}[k] \:]^{\top} \in \mathbb{R}^2$ is the control input and $k$ is the discrete-time index. In the following discussion, $A^d_i$ and $B^d_i$ are used to denote the state and input matrices respectively in \eqref{cont_dyn} for each agent. We assume each UAV starts at the initial position $ p^0_i =[p^{x_0}_i, p^{y_0}_i]^{\top} \in \mathbb{R}^2$ and has $ p^T_i =[p^{x_T}_i, p^{y_T}_i]^{\top} \in \mathbb{R}^2$ as its target position . We re-derive \eqref{cont_dyn} using the error vector $e_i[k]=z_{i}[k]-[p^{x_T}_i, \: p^{y_T}_i, \: 0,   \: 0 \:]^{\top}$. The error dynamics for each agent can be written as:
\begin{equation}
\label{error_dyn}
e_{i}[k+1]=A^d_{i}e_{i}[k]+B^d_{i}a_{i}[k]
\end{equation}

\subsubsection{Agent-wise optimization problem:}
Each agent $i \in I$ solves a receding horizon control problem of horizon length $H$. Let $u_i = [a_i[0] \ldots a_i[H-1]]^T_i$ be the decision vector of agent $i$ over the prediction horizon, where $u_i \in U_i \subseteq \mathbb{R}^{2H}$ denotes the presence of local constraints $U_i$ (to be defined in the sequel). Thus, each agent $i\in I$ solves:
\vspace{-4mm}

\begin{align}
\min_{u_i\in U_i}~ &J_i(u_i,\sigma(u)) \nonumber \\
\text{subject to }&
A_i u_i + \sum\limits_{j \in I \setminus \{i\}}{}A_j u_j \leq\sum\limits_{i \in I}b_i \label{dist_opt_eq}
\end{align}

where $j \in I \setminus \{i\}$ represents all agents except agent $i$ and  $J_i(u_i,\sigma(u))$ is the finite horizon cost which each agent $i \in I$ seeks to minimize. This cost is coupled via the aggregative variable $\sigma(u):= \frac{1}{N}\sum_{i \in I} \phi_i(u_i)$, which depends on all agents' decisions $u_i$, formed by the aggregation rule $\phi_i(\cdot)$. 

Let
$
p_o=[p_o^x,p_o^y]^\top
$
denote the common target position and 
$
d_o=
\left\|
\frac{1}{N}\sum_{z=1}^{N} p_z[H]-p_o
\right\|_2
$
the distance between the predicted swarm centroid at horizon step $H$ and $p_o$. We define the local cost function as follows: 
\begin{align}
& J_i(u_i,\sigma(u))= \sum_{l=0}^{H-1} e_i[l]^T Q_i e_i[l]
+ u_i[l]^T R_i u_i[l] \nonumber \\
& + \beta \, e_i[H]^T P_i e_i[H]
+ \frac{p_a}{N} d_o^2
\label{cost_fcn}
\end{align}
where the first two terms in \eqref{cost_fcn} represents the running cost with weighting matrices $Q_i\succ0$ and $R_i\succ0$, while $\beta P_i\succ0$ for third term defines the terminal penalty. The final term, weighted by $p_a/N$, penalizes the swarm-centroid deviation relative to individual target positions. The local constraint set $U_i$ captures the bounded UAV acceleration inputs:

\begin{align}
U_i=\{u_i:-a_i^{\max}\le a_i[k]\le a_i^{\max},\; k=0,\ldots,H-1\}
\label{set_Xi}
\end{align}


The agent-wise optimization is also subject to coupled constraints introduced for collision avoidance. This is defined for each pair of agents via the discrete-time control barrier function. For $i,j \in I$ with $i \neq j$, let $\delta p_{i,j}[k]=[p_{x_i}[k] \: p_{y_i}[k]]^T-[p_{x_j}[k] \: p_{y_j}[k]]^T$ and a set $S$, being the superlevel set of a map $ h^j_i:\mathcal{P} \subset \mathbb{R}^{2} \to \mathbb{R}$ 
\begin{equation*}
    S = \{\delta p_{i,j}[k] \in \mathcal{P} \subset \mathbb{R}^{n} : h^j_i(\delta p_{i,j}[k]) \geq 0\} .
\end{equation*}
The set $S$ denotes the safe set within which the agents' trajectories must remain to avoid collisions. To ensure collision avoidance:
\begin{flalign}
\label{cbf_ineq}
& (i)~ h^j_i(\delta p_{i,j}[0]) \geq 0, \nonumber \\
& (ii)~ \exists \: u_i[k], u_j[k] \text{ such that } \forall k \in \mathbb{N} \cup \{0\}, \\
&\hspace{0.6cm} h^j_i(\delta p_{i,j}[k+1])-h^j_i(\delta p_{i,j}[k]) \geq -\gamma_{\mathrm{cbf}} (h^j_i(\delta p_{i,j}[k])) \nonumber. &
\end{flalign}
The function $h^j_i:\mathcal{P} \rightarrow \mathbb{R}$ is said to be a discrete-time exponential CBF and the set $S$ is invariant, or safe, along the trajectories of \eqref{error_dyn} driven in closed loop by $u_i[k], u_j[k]$ in \eqref{cbf_ineq}, as defined in \cite{agrawal_discrete_2017}. The following candidate control barrier functions are used for $(i,j) \in I, i \neq j$: 
\begin{align}
h_i^j&(\delta p_{i,j}[k]) = \frac{\lvert p_{x_i}[k] - p_{x_j}[k] \rvert}{r_1} + \frac{\lvert p_{y_i}[k] - p_{y_j}[k] \rvert}{r_2} - 1 \label{cbf_absolute}\\
& = \frac{\lvert e_{i}^x[k] - e_{j}^x[k] + \delta p^d_{x_{i,j}} \rvert}{r_1} + \frac{ \lvert e_{i}^y[k] - e_{j}^y[k] + \delta p^d_{y_{i,j}}\rvert}{r_2} - 1 \nonumber.
\end{align}
The function $h_i^j(\delta p_{i,j}[k]) \geq 0$ guarantees no collision between UAVs $i$ and $j$ at time $k$. Parameters $r_1, r_2 > 0$ are the norm-1 unsafe radii, delimiting the UAV's body in the $x$ and $y$ directions. These conditions are then transformed into matrices $A_i$, $A_j$ and vectors $b_i$ defined in \eqref{dist_opt_eq}. For the full derivation and characterization of $A_i$, $A_j$ and $b_i$, we refer the reader to \cite{do_nascimento_game_2023}.

\subsubsection{Distributed optimization:}
Given our distributed safety requirements, the coupled constraints set the stage for a generalized games setting: typically, primal-dual iterative methods are then used, in which the Nash Equilibrium (NE) of the game is reached asymptotically. To enhance the real-time deployment potential, and effective peer-to-peer communication, we resort to a technique based on gradient tracking by \cite{trades_2024}. This technique is able to uncouple the cost and constraints by introducing two local variables on each agent, being able to track asymptotically the aggregative cost and coupled constraints in each agent. Different from \cite{do_nascimento_game_2023}, in this work we distribute the computation among different computer cores and study its high fidelity deployment numerically and in the hardware. For the full algorithm definitions and derivations, we refer the reader to \cite{trades_2024}. We highlight that the platform uses a modular distributed optimization scheme, and could be readily substituted by other algorithm of choice, such as the Alternating Direction Method of Multipliers (ADMM), a popular distributed optimization benchmark technique.

\subsection{Software Setup}
\label{subsec:software setup}

The framework was implemented using the MATLAB\textsuperscript{\textregistered} Parallel Computing Toolbox, where agents execute in parallel within an SPMD block and each computational engine corresponds to a worker. Algorithm~1 summarizes the software workflow.

For UAV prototyping, the double-integrator dynamics in \eqref{cont_dyn} were used for initial testing. The distributed position-swapping routine from \cite{do_nascimento_game_2023,trades_2024} was executed within the SPMD framework, where each worker represented a UAV agent maintaining local states, solving the distributed optimization problem, and exchanging neighbour information.

\begin{figure}[!ht]
\centering
\begin{minipage}{\columnwidth}
\small
\noindent\textbf{Algorithm-1 \newline Pseudocode for distributed control using SPMD}\\[-0.2em]
\rule{\linewidth}{0.4pt}

\begin{lstlisting} [basicstyle=\small\ttfamily]
% initialize variables e.g. numUAVs,maximum number of iteration steps, etc.
setup parameters
spmd (numUAVs)
  % Each worker represents a UAV agent
  id = spmdIndex;  % spmdIndex is the agent's number. 
  % 0. initialize state and define neighbours
  for k = 1:steps     % iterations 
      % 1. Communicate 
           Exchange information with neighbours 
           using spmdSendReceive 
      % 2. Compute local control action 
           using chosen iterative algorithm  
      % 3. Sense and Act 
           Measure state and apply action 
           to the system i.e. simple model/
           high fidelity model/ real hardware
      % 4. Synchronization between agents 
          (Optional step- depends on 
          type of routine
      % 5. Log data or send to base station
  end
end
% post processing and visualizations
\end{lstlisting}

\vspace{-0.2cm}
\rule{\linewidth}{0.4pt}
\label{fig:spmd_algorithm}
\end{minipage}
\end{figure}

For high-fidelity testing, the distributed routine was integrated with a Simulink model adapted from \cite{kimmodel}, based on the non-linear Crazyflie~2.0 dynamics identified in \cite{forster_cfmodel_2015}. The model includes modified low-level PID controllers. Additional interfaces were added to integrate with the SPMD framework. Using the MATLAB 2024a ``Simulation object'' feature%
\footnote{\url{https://tinyurl.com/simulink-simulation}}, each worker executes its local UAV simulation while exchanging distributed control information with neighbouring workers. This modular structure preserves the communication, synchronization, and logging routines while routing agent dynamics through either simplified, high-fidelity, or hardware-integrated models.
The complete distributed control framework codebase is publicly available at Github%
\footnote{\url{https://tinyurl.com/smart-city-dcf}}, including utilities for high-fidelity simulation, hardware experiments, data collection, and visualization.

\subsection{Hardware Setup and Integration}

\begin{figure}[!t]
\centering
\includegraphics[
    width=\columnwidth,
    trim={1cm 3cm 12cm 1cm},
    clip
]{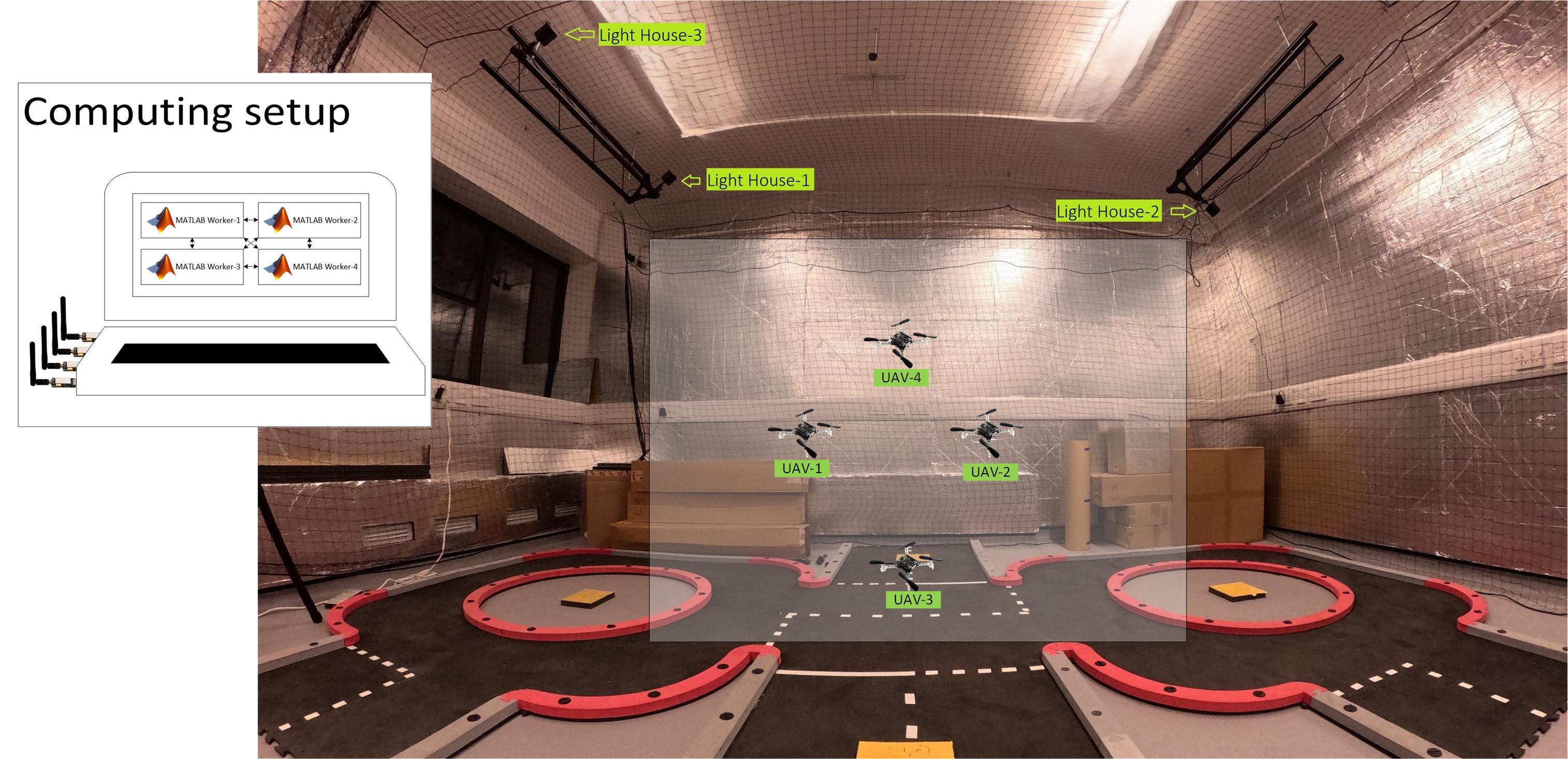}
\caption{Hardware setup for distributed control of multi-UAV systems using a single multi-core computer.}
\label{fig:hardware_setup}
\end{figure}

For UAV experiments, Crazyflie~2.1 drones were used which comes with its Python API%
\footnote{\url{https://github.com/bitcraze/crazyflie-lib-python}} for wireless communication and telemetry acquisition. The hardware setup is shown in Fig.~\ref{fig:hardware_setup}. Localization is provided by the Lighthouse-V2 infrared motion-tracking system, where onboard photodiodes detect laser sweeps from multiple base stations to triangulate the UAV pose with approximately 1\,cm accuracy \citep{taffanel_lighthouse_2021}. This proved sufficient for our distributed control experiments.

Communication with the UAVs is established using Crazyradio%
\footnote{\url{https://www.bitcraze.io/products/crazyradio-pa/}} USB dongles operating in the 2.4\,GHz ISM band. Although a single dongle can support multiple UAVs, one Crazyradio was assigned per agent to reduce latency and communication contention. The interface is managed through the Crazyflie Python API and integrated into MATLAB through Python calls. The radios support data rates up to 2\,Mbps and the Lighthouse-V2 system operates at approximately 30\,Hz, which were sufficient for waypoint updates to the onboard controllers.

The full command chain works as follows: after each distributed optimization step (Algorithm~1, Step~2), every worker transmits planar velocity references to its assigned Crazyflie. Using Lighthouse-V2 and onboard sensor measurements, the UAV estimates its state via an extended Kalman filter. The onboard PID velocity and attitude controllers then convert these references into motor PWM commands to track references. Updated state estimates are transmitted back to workstation through the Crazyradio interface for the next distributed control iteration.

\begin{figure*}[!t]
\centering

\begin{subfigure}{0.32\textwidth}
    \centering
    \includegraphics[width=\linewidth]{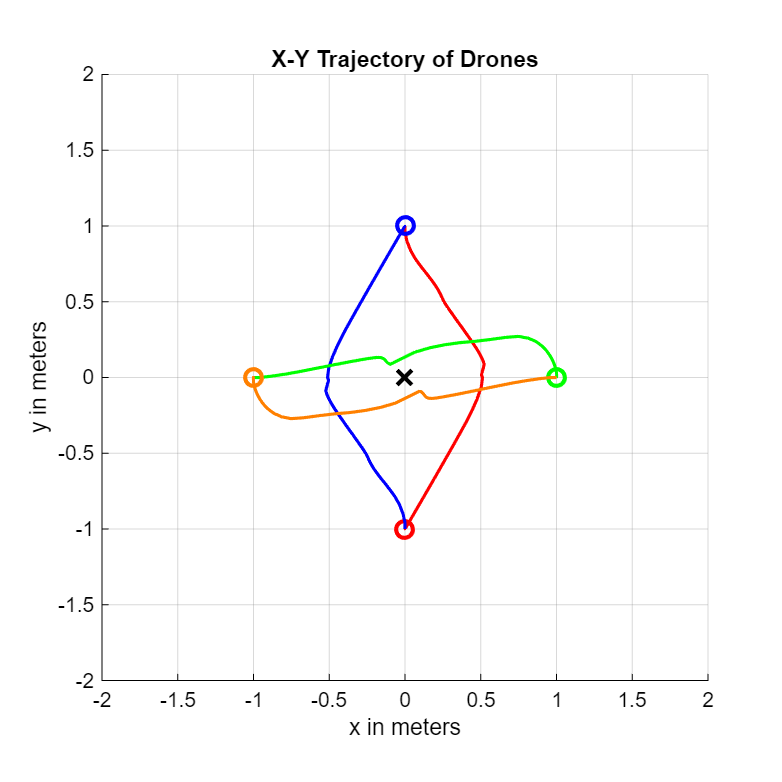}
    \label{fig:traj_integrator}
\end{subfigure}\hspace{-2mm}
\begin{subfigure}{0.32\textwidth}
    \centering
    \includegraphics[width=\linewidth]{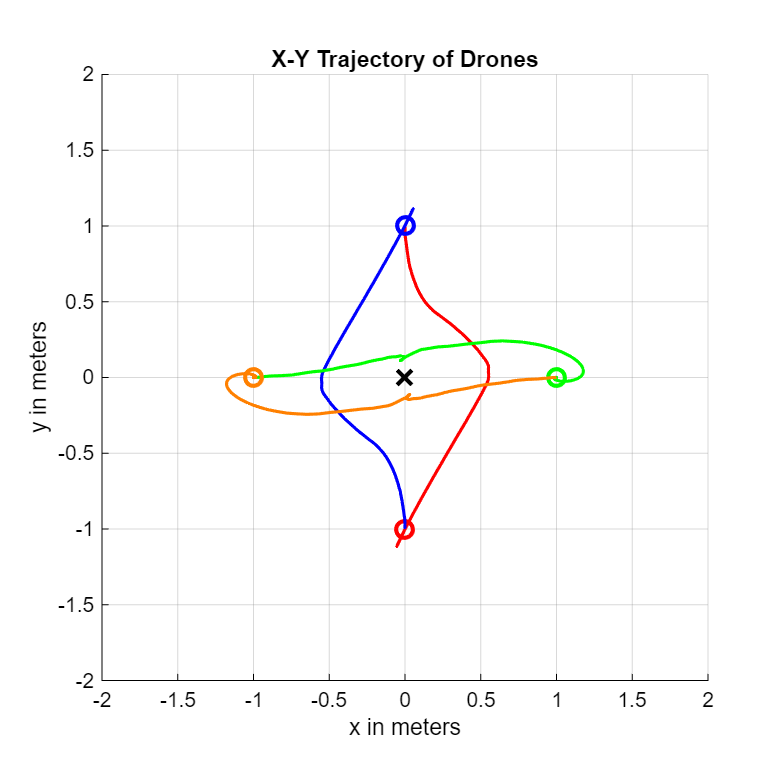}
    \label{fig:traj_hf}
\end{subfigure}\hspace{-2mm}
\begin{subfigure}{0.32\textwidth}
    \centering
    \includegraphics[width=\linewidth]{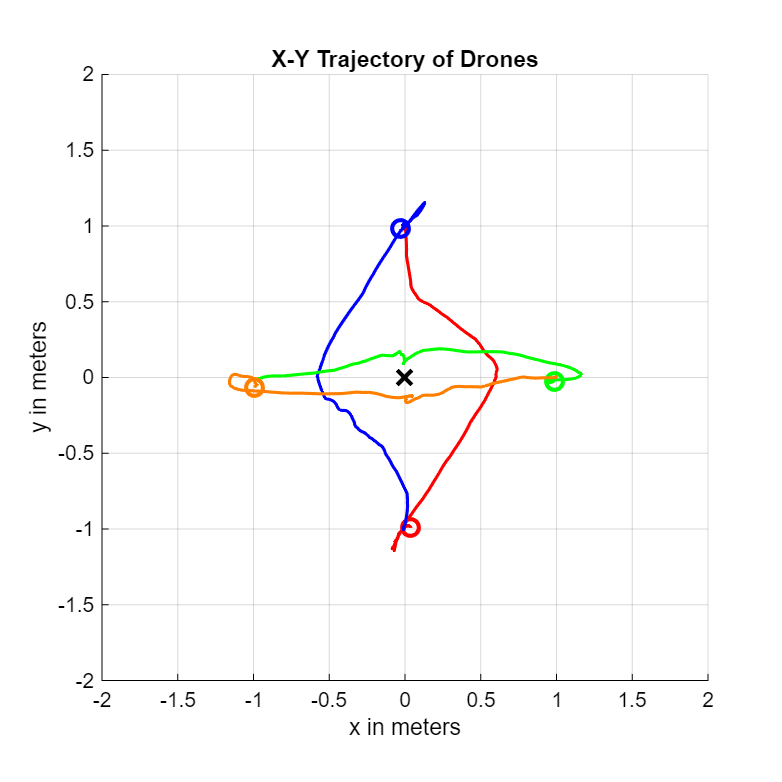}
    \label{fig:traj_uavs}
\end{subfigure}
\vspace{-5mm}
\caption{Trajectories for (left) double integrator, (center) high-fidelity, and (right) UAV experiments.}
\label{fig:traj_comparison}
\end{figure*}


\section{Results and Discussion}
\label{sec:Results}
We evaluate a four-UAV position-swapping task using a synchronous distributed controller over a fully connected communication network. Trajectories are compared across the double-integrator model, the Crazyflie high-fidelity model, and Crazyflie~2.1 hardware using the Lighthouse-V2 localization system. The initial positions are $p^0_{1}=[0,1]^T$, $p^0_{2}=[0,-1]^T$, $p^0_{3}=[1,0]^T$, and $p^0_{4}=[-1,0]^T$, with targets defined as $p^T_{1}=p^0_{2}$, $p^T_{2}=p^0_{1}$, $p^T_{3}=p^0_{4}$, and $p^T_{4}=p^0_{3}$. The common tracking point is $p_o=[0,0]^T$ with fleet weight $p_a=1$.

The sample time is $t_s=0.2\,\mathrm{s}$ and the acceleration bounds are $a^{\max}=2\,\mathrm{m/s^2}$ in both planar directions. The optimization parameters are $H=3$, $Q_i=\mathrm{diag}(5,5,5,5)$, $R_i=\mathrm{diag}(2,2,2,2)$, and $P_i$ equal to the Riccati solution of the unconstrained LQR problem without the aggregative term. Additionally, $\beta=1.5$, $\gamma_{\mathrm{cbf}}=0.1$, and safety radii $r_1=0.5\,\mathrm{m}$ and $r_2=0.25\,\mathrm{m}$.

Figure~\ref{fig:traj_comparison} compares the controller across all platform levels. Circles mark the trajectory endpoints, with the opposite ends indicating the UAV starting positions. The similar trajectories suggest that the double-integrator model is a reasonable motion-planning approximation, as low-level PIDs quickly correct tracking errors. The high-fidelity model, however, captures overshoot also observed in hardware, making it useful for parameter tuning. Early termination, needed for real-time deployment since the algorithm reaches the NE only asymptotically, weakens safety guarantees. The stopping criterion requires the maximum of the primal and dual differences between consecutive iterates, e.g., p and p+1, to fall below a heuristic threshold. Despite this and the model mismatch, actual collisions were rarely observed. The high-fidelity model shows two agents briefly entering each other's safety region and overshoot when reaching the target point, later confirmed in hardware but absent in the point-mass model. No collisions occurred, partly due to the generous CBF safety radii. A recorded demonstration is available online\footnote{\url{https://youtu.be/xZksNosFYro?si=wq258DJx4m99w_6R}}.

\begin{figure}[!b]
\begin{center}
\includegraphics[
    width=0.85\columnwidth,
    trim={0cm 1cm 0cm 0.5cm},
    clip
]{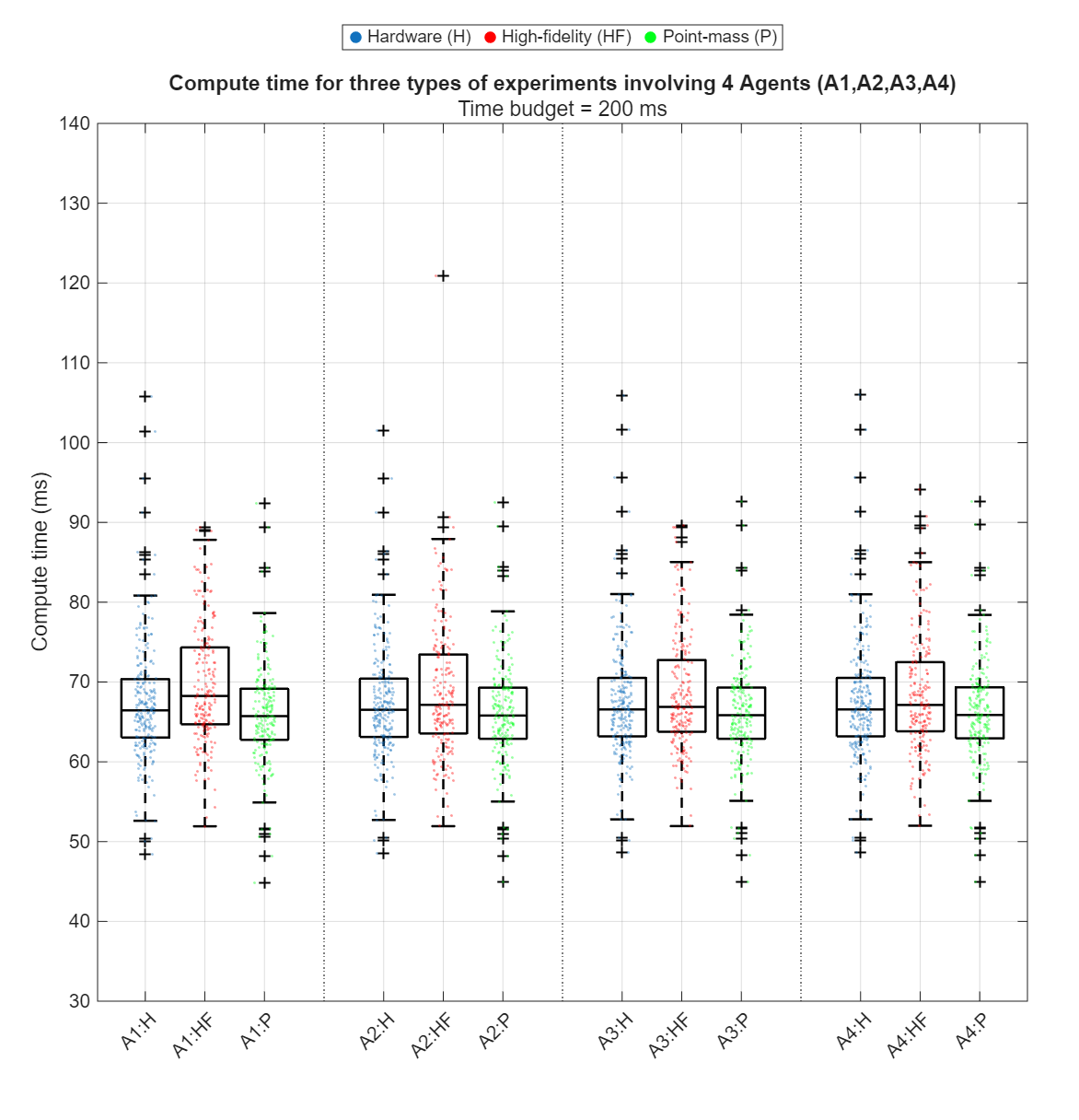}
\caption{Computation time distribution for four agents.}
\label{fig:iteration_time}
\end{center}
\end{figure}

Figure~\ref{fig:iteration_time} summarizes the computational performance of the distributed MPC controller across all framework levels. The boxplots show computation times measured around Step~2 of Algorithm~1 for the point-mass model, high-fidelity Crazyflie simulator, and Crazyflie~2.1 hardware. After removing one extreme outlier using a $10\times\mathrm{IQR}$ criterion, all platforms exhibit similar distributions with medians well below the $200\,\mathrm{ms}$ real-time budget. The hardware results closely match the point-mass model in both median and variability, supporting the use of simplified models for early-stage algorithm testing. The high-fidelity model is consistently slower since each worker simulates its local UAV dynamics. Overall, the results demonstrate real-time feasibility of the distributed MPC scheme across all testing levels.

\section{Conclusion}
\label{sec:conclusion}
This paper presented a prototyping framework for distributed control of multi-robot systems, combining parallel computing with a modular structure that supports an expanding library of simulation models and hardware APIs. Integration with Crazyflie quadrotor hardware demonstrated efficient single-workstation validation of distributed algorithms without complex networking infrastructure or dedicated multi-robot testbeds.
The approach has limitations: wireless interference, radio contention, and physical timing jitter cannot be fully reproduced in single-host emulation and must be assessed on real hardware. Nevertheless, the framework provides a fast, low-cost, and flexible environment for iterative development, reducing the effort required for performance testing. Future work will expand the model and hardware libraries, enabling broader comparisons across fidelity levels and robotic platforms. The framework may also be used to study communication effects and adversarial disturbances, supporting robustness analysis in multi-robot distributed control.
\vspace{-0.15cm}

\bibliography{ifacconf}  

\end{document}